\documentclass[10pt,twocolumn,letterpaper]{article}

\usepackage[pagenumbers]{iccv} 

\usepackage{makecell}
\usepackage[accsupp]{axessibility}  

\definecolor{iccvblue}{rgb}{0.21,0.49,0.74}
\usepackage[pagebackref,breaklinks,colorlinks]{hyperref}
\usepackage{multirow}
\usepackage{siunitx}
\usepackage{bbm}
\usepackage{float}
\def\paperID{*****} 
\def\confName{ICCV}
\def\confYear{2025}

\title{Describe Anything Model for Visual Question Answering on Text-rich Images}

\author{
Yen-Linh Vu\textsuperscript{1}\thanks{Equal contribution.}\quad
Dinh-Thang Duong\textsuperscript{1}\footnotemark[1]\quad
Truong-Binh Duong\textsuperscript{1}\quad
Anh-Khoi Nguyen\textsuperscript{1}\quad 
Thanh-Huy Nguyen\textsuperscript{2} \quad \\
Le Thien Phuc Nguyen\textsuperscript{3} \quad
Jianhua Xing\textsuperscript{4} \quad
Xingjian Li\textsuperscript{2} \quad  
Tianyang Wang\textsuperscript{5} \quad
Ulas Bagci\textsuperscript{6} \quad
Min Xu\textsuperscript{2}\thanks{Corresponding author: \texttt{mxu1@cs.cmu.edu}.\\
We thank AI VIETNAM for facilitating computational resources and financial support for this paper.}\\
\textsuperscript{1}AI VIETNAM Lab, Vietnam \\
\textsuperscript{2}Carnegie Mellon University, USA\\
\textsuperscript{3}University of Wisconsin - Madison, USA\\
\textsuperscript{4}University of Pittsburgh, USA\\
\textsuperscript{5}University of Alabama at Birmingham, USA\\
\textsuperscript{6}Northwestern University, USA
}

\begin{document}
\maketitle
\begin{abstract}
Recent progress has been made in region-aware vision-language modeling, particularly with the emergence of the Describe Anything Model (DAM). DAM is capable of generating detailed descriptions of any specific image areas or objects without the need for additional localized image-text alignment supervision. We hypothesize that such region-level descriptive capability is beneficial for the task of Visual Question Answering (VQA), especially in challenging scenarios involving images with dense text. In such settings, the fine-grained extraction of textual information is crucial to producing correct answers. Motivated by this, we introduce DAM-QA, a framework with a tailored evaluation protocol, developed to investigate and harness the region-aware capabilities from DAM for the text-rich VQA problem that requires reasoning over text-based information within images. DAM-QA incorporates a mechanism that aggregates answers from multiple regional views of image content, enabling more effective identification of evidence that may be tied to text-related elements. Experiments on six VQA benchmarks show that our approach consistently outperforms the baseline DAM, with a notable 7+ point gain on DocVQA. DAM-QA also achieves the best overall performance among region-aware models with fewer parameters, significantly narrowing the gap with strong generalist VLMs. These results highlight the potential of DAM-like models for text-rich and broader VQA tasks when paired with efficient usage and integration strategies. Our code is publicly available at \url{https://github.com/Linvyl/DAM-QA.git}.
\end{abstract}
    
\section{Introduction}
\label{sec:intro}

\begin{figure}
    \centering
    \includegraphics[width=\linewidth]{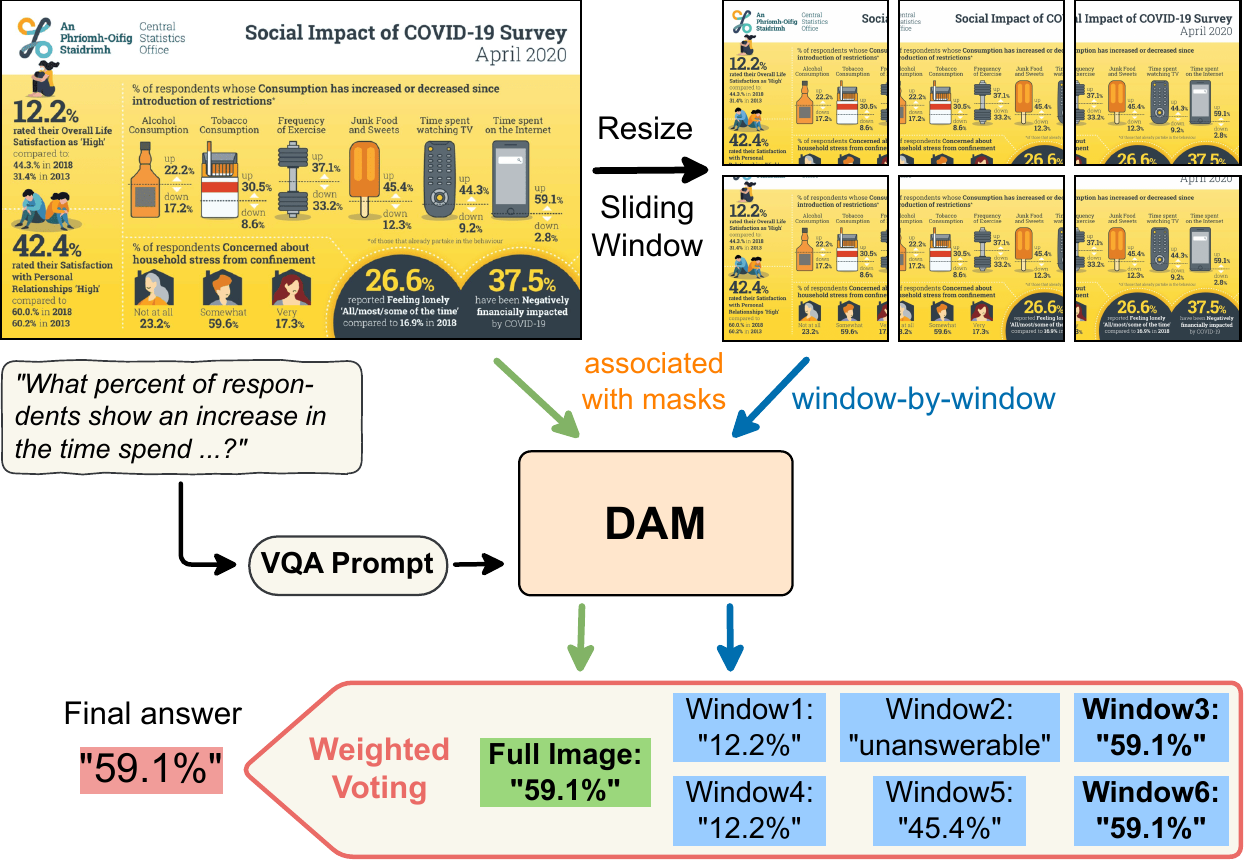}
    \caption{DAM-QA handles text-rich VQA by applying DAM to both global and local image views, capturing focused regions through a sliding-window process. The resulting answers are aggregated through voting to enhance overall prediction accuracy.}
    \label{fig:abstract_TextualDAM}
\end{figure}

Text-rich Visual Question Answering (VQA) is a challenging multimodal task that demands a system to answer natural language questions based on visuals with considerable textual content, such as documents, infographics, and charts~\cite{mathew2021docvqa, mathew2022infographicvqa, masry2022chartqa}. Solving this problem entails not only recognizing and reading text in various styles and layouts but also comprehending how textual pieces interact with the surrounding visual context. The intricacy and diversity of visual-textual interactions in such images present significant problems for current approaches~\cite{powalski2021going}.

Traditionally, dense text VQA methods rely heavily on Optical Character Recognition (OCR) to extract embedded text, followed by fusion modules that integrate visual and textual features~\cite{9710059,10.1145/3394486.3403172,xu-etal-2021-layoutlmv2,nguyen2025enhancing,10656994}. However, these pipelines are often brittle, sensitive to OCR errors and reliant on complex post-processing. Similar to many works that leverage global-local visual mechanism~\cite{ngo2024dual,tran2025igl,tran2025mlg2net}, end-to-end Vision-Language Models (VLMs) have emerged as unified alternatives that jointly encode visual and textual inputs~\cite{bai2025qwen2, zhu2025internvl3, abouelenin2025phi}. While effective, many VLMs primarily focus on global image-text alignment, limiting their ability to capture fine-grained information in dense or localized regions.

Region-aware VLMs have gained attention by incorporating localized supervision or attention, showing promise in tasks like referring expression comprehension and grounded question answering~\cite{chen2023shikra, guo2024regiongpt, cai2024vip, zhang2025gpt4roi}. Notably, the Describe Anything Model (DAM)~\cite{lian2025describe} supports diverse region formats (points, masks, boxes) and excels at generating fine-grained region-level descriptions with minimal supervision. While DAM has proven effective for region-level description, its potential in complex reasoning tasks like dense text VQA remains underexplored. We posit that DAM’s flexible region-aware modeling makes it a strong candidate for handling text-rich visual reasoning tasks.

To this end, we introduce DAM-QA, a framework that extends DAM’s capabilities to text-rich VQA. Inspired by prior work on localized visual understanding~\cite{kant2020spatially, lin2022revive, Appalaraju_Tang_Dong_Sankaran_Zhou_Manmatha_2024, liu2024textmonkey}, DAM-QA applies a sliding-window approach alongside full-image inference to extract regional patches, enabling multi-scale context modeling. These multiple views are fused via a voting-based strategy to enhance robustness against incomplete or ambiguous inputs. A custom evaluation protocol ensures fair assessment of DAM's VQA ability without modifying its architecture.

We assess DAM-QA against six available VQA benchmarks, which include both document-centric and general-purpose datasets. While DAM-QA does not yet equal the performance of generalist foundation models, it beats the baseline DAM and other region-aware VLMs on most benchmarks, including a significant increase of over 7 points on DocVQA~\cite{mathew2021docvqa} over the original DAM model. These findings demonstrate that modifying region-aware models, such as DAM, has the potential to enhance performance in VQA tasks, particularly those that involve fine-grained reasoning over localized textual information.

Our contributions can be summarized as follows:

\begin{itemize}
\item We provide a systematic investigation into applying region-aware VLMs to text-rich VQA tasks, using the Describe Anything Model (DAM) as a representative case.
\item We introduce DAM-QA, a simple yet effective framework combining region-level inference and answer aggregation without requiring additional supervision.
\item We report comprehensive results on six VQA benchmarks, showing improved average performance over the DAM baseline and demonstrating the value of region-aware representations.
\end{itemize}

\section{Related Works}
\label{sec:related_works}

\subsection{Vision-Language Models}

Vision-Language Models (VLMs) support tasks like captioning, retrieval, grounding, and Visual Question Answering (VQA) by learning joint image-text representations using Transformer architectures trained with masked or contrastive objectives. Masked VLMs align visual regions and text via masked prediction. Early models~\cite{li2019visualbert, lu2019vilbert} fused object features with word tokens under BERT-style pretraining. Later models~\cite{li2020oscar} introduced richer image-text objectives, setting new benchmarks in VQA. Advances such as~\cite{huang2021seeing} dropped region proposals via masked visual modeling, while~\cite{li2020unimo} improved robustness and scale. InternVL3~\cite{zhu2025internvl3} added native multimodal pretraining and variable visual encoding, rivaling proprietary models like Gemini 2.5~\cite{gemini25}. Autoregressive VLMs connect vision encoders to frozen LLMs via instruction tuning. The BLIP family~\cite{li2023blip, li2022blip} introduced Q-Formers for vision-to-language alignment. Models like~\cite{alayrac2022flamingo} extend this to few-shot generation, while~\cite{Dai2023InstructBLIPTG} use GPT-4-style dialogue to enhance instruction following. Recent open models~\cite{abouelenin2025phi, dong2025scalable} show strong VQA and OCR performance, offering transparency and comparability, unlike proprietary models such as Gemini 2.5 Flash~\cite{gemini25}. Despite these advances, most VLMs operate on whole-image inputs and lack explicit mechanisms for region-level grounding, especially in text-heavy images. This motivates region-aware VLMs, which aim to better align spatially localized visual and textual features, as explored in the next section.

\begin{figure*}[t]
  \centering
  \includegraphics[width=0.98\textwidth]{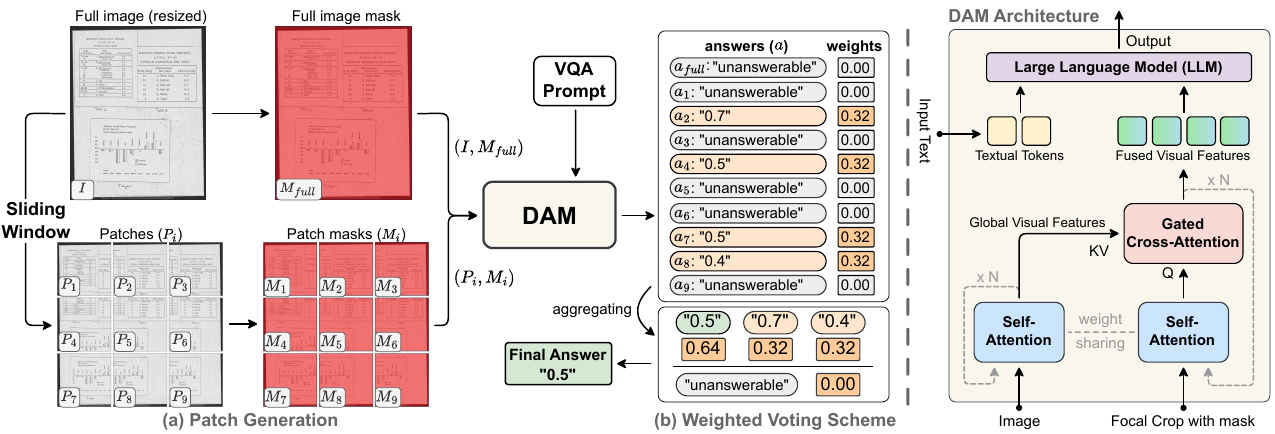}
\caption{\textbf{Overview of the proposed DAM-QA framework}
  \textbf{(a) Patch Generation} (\ref{sec:patch_generation}): An input image is processed with a sliding window to generate a full-image view alongside multiple overlapping local patches.
  \textbf{(b) Weighted Voting Scheme} (\ref{sec:weighted_voting_scheme}): All views are fed into the DAM model with a unified VQA prompt. The resulting answers are aggregated via a weighted voting system, where ``unanswerable'' predictions are assigned a weight of 0, while the full-image prediction receives a weight of 1 and patch-level predictions are weighted by their area. The final answer is determined by the highest total score.
}
  \label{fig:pipeline}
\end{figure*}




\subsection{Region-aware Vision-Language Models}

Most existing VLMs operate at image-level granularity, lacking fine-grained spatial grounding capabilities. Yuan et al.~\cite{zhang2025gpt4roi} highlight the shortage of region-level training data as a key limitation.
To address this, region-aware VLMs try to integrate spatial prompts (such as points, boxes, masks, or scribbles) or generate region-specific outputs. These models~\cite{peng2023kosmos, chen2023shikra, you2023ferret, zhang2025gpt4roi, yuan2024osprey, cai2024vip, lian2025describe, lin2025perceive, xiao2025describe} tackle tasks such as region captioning, referring expression comprehension, spatial reasoning, and region-based question answering. Notable region-aware methods include KOSMOS~\cite{peng2023kosmos}, which encodes spatial references via markdown links; Shikra~\cite{chen2023shikra}, which expresses coordinates in natural text; Ferret~\cite{you2023ferret}, which fuses spatial tokens with visual features; GPT4RoI~\cite{zhang2025gpt4roi}, which introduces spatial instruction tuning; and Osprey~\cite{yuan2024osprey}, which applies pixel-level attention using mask-text pairs. More recently, ViP-LLaVA~\cite{cai2024vip} integrates visual overlays (e.g., arrows, boxes, scribbles) into a CLIP-based pipeline.
Describe Anything Model (DAM)~\cite{lian2025describe} is designed explicitly for localized captioning that employs focal prompts to capture high-resolution details and a localized backbone to retain global context.

\section{Methodology}
\label{sec:methods}

In this section, we first describe the task formulation (Section~\ref{sec:task_definition}) and our baseline zero-shot Visual Question Answering (VQA) procedure, which performs full-image mask inference with a single call to the Describe Anything Model (DAM) (Section~\ref{sec:baseline_dam}). We then introduce the details of our proposed framework, DAM-QA, including its sliding-window extension designed to capture small text regions that the global view may miss (Section~\ref{sec:sliding}), and a prompt strategy crafted to enable DAM for VQA scenarios (Section~\ref{sec:prompt}). At each step, we explain the reasoning behind parameter choices and provide the essential mathematical formulation. Figure~\ref{fig:pipeline} illustrates the overall workflow.

\subsection{Visual Question Answering (VQA) Task}
\label{sec:task_definition}

The VQA task requires a model to answer a question based on a given image. This work focuses on a challenging subdomain of the task, namely text-centric VQA. In this setting, accurately answering the question depends on the model's ability not only to perceive visual elements, but also to read, understand, and reason over textual information within the image, such as in documents, infographics, and charts. Nevertheless, to assess the broader applicability of our method, we also evaluate it on a general-purpose benchmark, VQAv2 \cite{goyal2017vqav2}, which involves more diverse question types beyond text-based reasoning. Despite our specific focus on text-rich scenarios, the problem can be formalized as a general VQA task. Given an input image $I$ and a question $Q$, the objective is to learn a function $f$ that predicts the answer $\hat{a} = f(I, Q)$. The predicted answer $\hat{a}$ is then evaluated against a ground truth, which can be a single reference answer $a$ or a set of $m$ acceptable answers $A = \{a_1, a_2, \ldots, a_m\}$, depending on the benchmark.


\subsection{Baseline DAM Inference for Zero‐shot VQA}
\label{sec:baseline_dam}
The Describe Anything Model (DAM)~\cite{lian2025describe} is a region‐aware vision-language system that takes as input an RGB image \(I\in\mathbb{R}^{W\times H\times 3}\) (where \(W\) and \(H\) are the image width and height in pixels) and a binary mask \(M\in\{0,1\}^{W\times H}\), with \(M(x,y)=1\) indicating that the pixel at coordinates \((x,y)\) belongs to the region of interest. For zero‐shot VQA, we construct the full‐image mask $M_{\mathrm{full}}(x,y)=1$ for all $(x,y)\in[0,W)\times[0,H)$, which directs DAM to attend to every pixel in the image.
Let \(\text{prompt}\) be the unified VQA prompt defined in Section~\ref{sec:prompt}.  We then invoke DAM exactly once:
\begin{equation}
\hat a_{\mathrm{full}}
\;=\;
\mathrm{DAM}\bigl(I,\;M_{\mathrm{full}},\;\text{prompt}\bigr).
\label{eq:baseline_call}
\end{equation}

Here, \(\hat a_{\mathrm{full}}\) is the prediction produced by DAM with \(M_{\mathrm{full}}\).  We remove any leading or trailing whitespace but apply no further filtering.  This single‐call procedure constitutes our baseline zero‐shot VQA inference across all datasets.

\subsection{Sliding Window Extension}
\label{sec:sliding}
To effectively capture multi-scale textual cues within text-rich images, we adopt a sliding-window extension motivated by recent works in high-resolution vision-language understanding, such as \cite{10658022, huang2025hires}. These methods demonstrate that dividing large or high-density visual inputs into overlapping local views (each processed independently) can significantly enhance fine-grained reasoning, especially when leveraging voting or fusion mechanisms for final prediction.

\subsubsection{Patch Generation}
\label{sec:patch_generation}
Each image is resized so that its longest side equals 1024 pixels, preserving aspect ratio. The resulting dimensions \(W\le1024\) and \(H\le1024\) define the grid for patch extraction. Square patches of size \(512\times512\) pixels are extracted in a sliding-window fashion with a stride of $256$‐pixels to ensure fifty-percent overlap (see Table~\ref{tab:ablation_patch}). Formally, we collect all top-left coordinates \(\{(x_i,y_i)\}_{i=1}^{N}\) satisfying
\[
0 \,\le\, x_i \,\le\, W - 512,
\qquad
0 \,\le\, y_i \,\le\, H - 512,
\]
are enumerated; any residual patches along the right or bottom edges are also added, and duplicate coordinates are removed to yield \(K\) distinct regions. If \(W<512\) or \(H<512\), then \(K\) is set to 1, with the full image treated as a single patch. The \(i\)th crop is denoted by \(P_i\) and its corresponding full-foreground mask by \(M_i\).

\subsubsection{Weighted Voting Scheme}
\label{sec:weighted_voting_scheme}
After obtaining \(K\) overlapping patches \(P_{1},\dots,P_{K}\) (with masks \(M_{1},\dots,M_{K}\)) alongside the full‐image mask \(M_{\mathrm{full}}\), each view is submitted to DAM under the unified VQA prompt.  The resulting predictions are denoted as
\begin{align}
  \hat a_{\mathrm{full}} 
    &= \mathrm{DAM}\bigl(I,\,M_{\mathrm{full}},\,\text{prompt}\bigr),
  \label{eq:full_pred}\\
  \hat a_{i} 
    &= \mathrm{DAM}\bigl(P_{i},\,M_{i},\,\text{prompt}\bigr),
    \quad i=1,\dots,K.
  \label{eq:patch_pred}
\end{align}

We assign the full‐image prediction a fixed vote weight of one and each patch prediction \(\hat a_{i}\) a weight
\begin{equation}
\alpha_{i}
=
\begin{cases}
\displaystyle
\frac{\mathrm{area}(P_{i})}{W\,H}, 
& \hat a_{i}\neq \text{``unanswerable''},\\[1ex]
0,
& \text{otherwise},
\end{cases}
\label{eq:alpha_i}
\end{equation}
where \(\mathrm{area}(P_{i})\) is the pixel area of patch \(P_{i}\) and \(W\,H\) is the total pixel count of the resized image.

To count votes for any candidate answer \(a\), we introduce the equality indicator
\begin{equation}
\mathbbm{1}[a=\hat a]
=
\begin{cases}
1, & \text{if \(a\) exactly matches the prediction \(\hat a\)},\\[0.5ex]
0, & \text{otherwise}.
\end{cases}
\label{eq:indicator}
\end{equation}

Votes are then accumulated as
\begin{equation}
\mathrm{votes}(a)
= \mathbbm{1}[a=\hat a_{\mathrm{full}}]
\;+\;\sum_{i=1}^{K}\alpha_{i}\;\mathbbm{1}[a=\hat a_{i}].
\label{eq:weighted_votes}
\end{equation}

If \(\sum_{i=1}^{K}\alpha_{i}=0\), meaning every patch prediction was “unanswerable,” the system falls back on the full‐image output and therefore returns whatever \(\hat a_{\mathrm{full}}\) predicted (which itself may be “unanswerable”).  Otherwise the answer \(a\) with the highest accumulated vote total is selected:
\begin{equation}
\hat a = \arg\max_{a}\,\mathrm{votes}(a).
\label{eq:final_answer}
\end{equation}

\begin{figure}
    \centering
    \includegraphics[width=\linewidth]{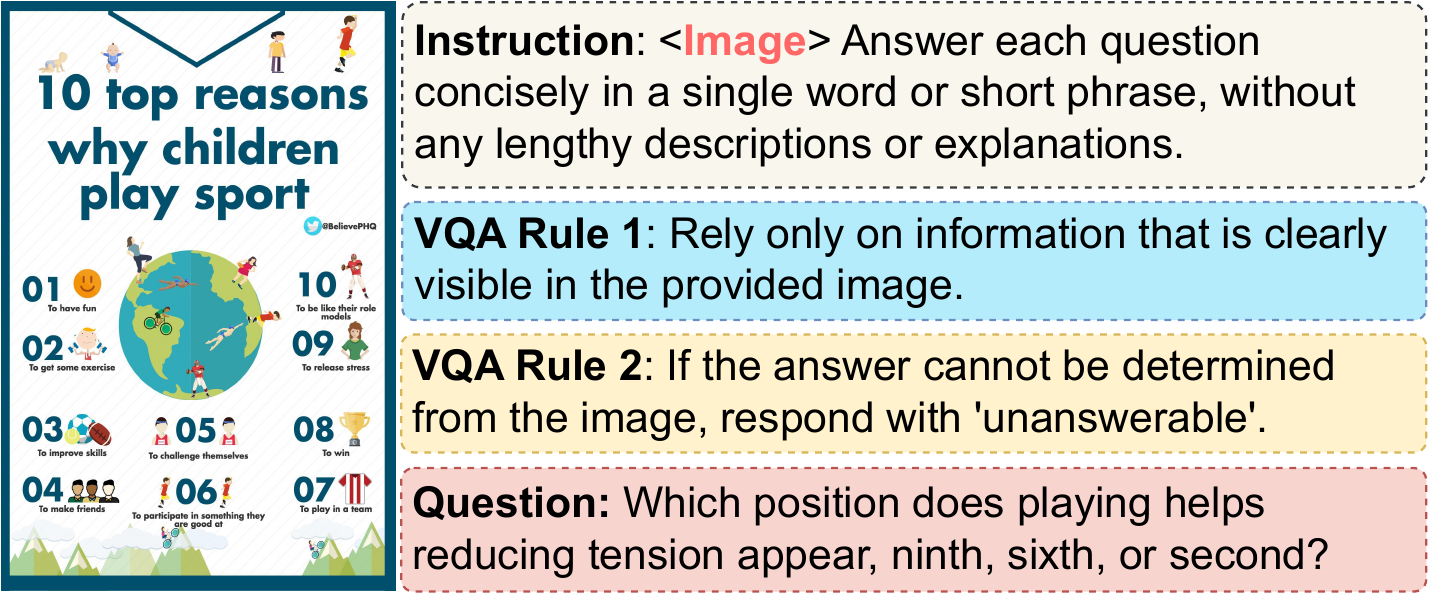}
    \caption{An illustration of the prompt construction used for evaluating Visual Question Answering models. The prompt consists of four components: a brief instruction, two rules that reflect and enforce the specialized requirements of the VQA task, and the input question.}
    \label{fig:prompt_vqa}
\end{figure}


\subsection{VQA Prompt Construction}
\label{sec:prompt}
All image inputs to DAM, whether from a global or local view, are paired with a unified prompt designed to enforce consistent behavior. In accordance with the requirements of the VQA task, the prompt is structured into four components. These include a general instruction, two VQA-specific rules, and the question itself, shown in Figure~\ref{fig:prompt_vqa}:

\begin{enumerate}
\item \textbf{General Instruction:} A brief directive asking the model to produce a concise answer in a single word or short phrase, without elaboration.
\item \textbf{VQA Rule 1 (Image-only Justification):} The model must ground its answer solely in the content visible within the image, disallowing the use of external knowledge. This supports faithful visual grounding, a core objective of the VQA task \cite{antol2015vqa, goyal2017vqav2}.
\item \textbf{VQA Rule 2 (Abstention for Insufficient Evidence):} The model is instructed to reply with “unanswerable” if the image alone does not contain sufficient information. This is aligned with benchmarks such as TextVQA \cite{singh2019textvqa} and ChartQAPro \cite{masry2025chartqapro}, which favor abstention over hallucination.
\item \textbf{Question:} The input query, which may be asked about any part of the image content.
\end{enumerate}

\section{Experimental Results}
\subsection{Datasets}
To evaluate performance on VQA tasks where accurate answers are heavily dependent on understanding the visual text, a suite of specialized datasets is employed: DocVQA, InfographicVQA, TextVQA, ChartQA, ChartQAPro (with conversational samples excluded), and VQAv2. For VQAv2, the evaluation adheres to the protocol of~\cite{jiang2024multi} by using the ``rest-val'' subset, comprising approximately 5,000 questions from the validation set. A detailed summary of these benchmarks, including their sources, core challenges, and data statistics, is provided in Table~\ref{tab:dataset_summary}.

\begin{table*}[!ht]
\centering
\resizebox{\textwidth}{!}{%
\begin{tabular}{@{}lp{4.5cm}p{6.6cm}p{1.4cm}p{3.55cm}@{}}
\toprule
\textbf{Dataset} & \makecell[c]{\textbf{Image Source \& Type}} & \makecell[c]{\textbf{Core Task \& Challenge}} & \textbf{\makecell{Images /\\QA Pairs}} & \makecell[c]{\textbf{Answer Type}} \\
\midrule
\makecell[lt]{DocVQA \\ \cite{mathew2021docvqa}} & Scanned documents from the UCSF Industry Documents Library. & Answering questions requires both reading the text and understanding the structure and layout of the document. & \makecell[ct]{12,767 /\\50,000} & Mostly Extractive (answers are verbatim spans from the document). \\
\addlinespace[0.2em]
\makecell[lt]{Info-\\graphic-\\VQA~\cite{mathew2022infographicvqa}} & Infographics from diverse web sources. Images combine text, graphics, and data visualizations. & Requires joint reasoning over layout, text, and graphical elements. Emphasizes elementary reasoning and basic arithmetic skills. & \makecell[ct]{5,485 /\\30,035} & Image-span, Question-span, Multi-span, and Non-extractive. \\
\addlinespace[0.2em]
\makecell[lt]{TextVQA \\ \cite{singh2019textvqa}} & Natural scenes from the Open Images dataset, selected from categories likely to contain text. & Bridges OCR and VQA by requiring models to read and reason about text in natural images to answer questions. & \makecell[ct]{28,408 /\\45,336} & Open-ended, typically handled via a copy mechanism from OCR tokens or a fixed vocabulary. \\
\addlinespace[0.2em]
\makecell[lt]{ChartQA \\ \cite{masry2022chartqa}} & Real-world charts from curated online data platform focused on social and economic topics. & Focuses on complex compositional questions that involve multiple logical and arithmetic operations over chart data. & \makecell[ct]{20,882 /\\32,719} & Open-vocabulary, often requiring generated numerical answers from calculations. \\
\addlinespace[0.2em]
\makecell[lt]{ChartQA-\\Pro \\\cite{masry2025chartqapro}} & Charts from highly diverse online sources. Features complex layouts like multi-chart images, infographics, and dashboards. & A more challenging benchmark designed to address the limitations of ChartQA with greater diversity and more complex, real-world question types. & \makecell[ct]{1,341 /\\1,948} & Multiple-Choice, Conversational, Hypothetical, Unanswerable, Fact-Checking. \\
\addlinespace[0.2em]
\makecell[lt]{VQAv2 \\ \cite{goyal2017vqav2}} & Natural images from the COCO dataset. & A general VQA task focused on visual reasoning. Balanced by collecting complementary image-question pairs to reduce language priors. & \makecell[ct]{204K /\\1,1M} & Open-ended (evaluated as classification over most frequent answers). \\
\bottomrule
\end{tabular}%
}
\caption{Summary of the VQA benchmark datasets used for evaluation. The datasets cover a diverse range of challenges, including scanned documents, infographics, natural images, and data visualizations. Each dataset presents unique demands, such as layout understanding, OCR integration, compositional reasoning, or open-domain visual question answering.}
\label{tab:dataset_summary}
\end{table*}

\subsection{Evaluation Metrics}
\label{sec:metrics}
Model performance is assessed using two types of metrics: benchmark-specific scores and a unified LLM-based semantic score.

\subsubsection*{Benchmark-specific Metrics}
To enable fair comparison with prior work, we use the official metric for each dataset: \textbf{ANLS} (Average Normalized Levenshtein Similarity) for DocVQA and InfographicVQA, \textbf{RAcc} (Relaxed Accuracy) for ChartQA and ChartQAPro, and the \textbf{VQAS} (VQA Score) for TextVQA and VQAv2. For brevity, we use these acronyms throughout the result tables.

\paragraph{Relaxed Accuracy (RAcc)}

RAcc is the primary metric for chart-based datasets, with its implementation varying by benchmark. The original ChartQA uses a simpler variant where numeric answers are correct within a $\pm 5\%$ relative error, while textual answers require an exact match. In contrast, ChartQAPro requires a strict exact match for question types with a constrained answer space and for all year-based answers to prevent false positives. For all other questions, it uses the same $\pm 5\%$ tolerance for numeric answers and ANLS for general text. This logic is formalized as:
\begin{equation} \label{eq:racc}
\text{score}(\hat{a}, a) =
\begin{cases}
    1 &~\parbox[t]{0.45\linewidth}{if the question is MCQ or Fact-Checking, and $\hat{a} = a$,} \\
    1 &~\parbox[t]{0.45\linewidth}{if $a$ is a year, and $\hat{a} = a$,} \\
    1 &~\parbox[t]{0.45\linewidth}{if $a$ is numeric (not a year) and $\frac{|\hat{a} - a|}{|a|} \le 0.05$,} \\
    \text{ANLS}(\hat{a}, a) & \text{otherwise}.
\end{cases}
\end{equation}
For questions with list-based answers, the score is the mean correctness of all items. The final metric is the average of these scores across the dataset.



\paragraph{Average Normalized Levenshtein Similarity (ANLS)}
ANLS accounts for minor OCR errors and slight wording variations by measuring the similarity between a predicted answer $\hat{a}$ and each ground truth $a_j$ using Normalized Levenshtein (NL) distance. The similarity function $s(\hat{a}, a_j)$ outputs a score in $[0, 1]$ and applies a cutoff threshold $\tau$ (typically 0.5):
\begin{equation} \label{eq:anls_similarity}
s(\hat{a}, a_j) = \begin{cases} 1 - \text{NL}(\hat{a}, a_j) & \text{if } \text{NL}(\hat{a}, a_j) < \tau, \\ 0 & \text{if } \text{NL}(\hat{a}, a_j) \ge \tau. \end{cases}
\end{equation}
The NL distance is the standard Levenshtein distance~\cite{levenshtein1966binary}, normalized by the length of the longer string. The ANLS score for a question is the maximum similarity across all ground truths, and final score is averaged over all samples.

\paragraph{VQA Score (VQAS)}
The standard VQA Score handles open-ended answer variability by first normalizing both predicted and reference answers. Let \(N_{\mathrm{agree}}\) be the number of reference answers that exactly match the normalized prediction. The per-question score is calculated as:
\begin{equation} \label{eq:vqas}
\text{score}(\hat{a}, A) = \min\left(N_{\mathrm{agree}}/3, 1.0\right).
\end{equation}
The final VQA Score is the mean of these per-question scores across all samples.
\begin{figure}[h!]
\centering
\includegraphics[width=\linewidth]{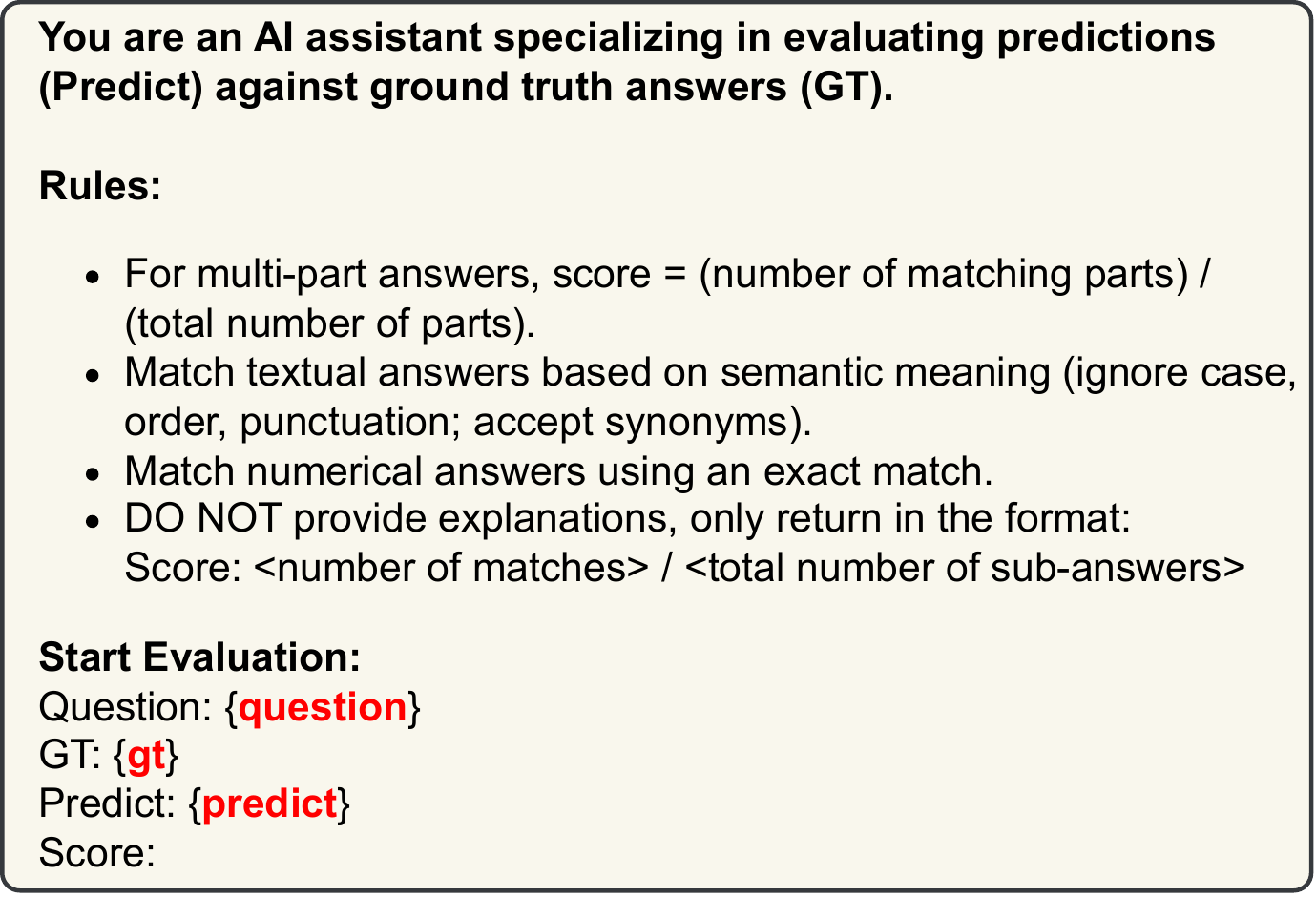}
\caption{LLM judge prompt template with placeholders (in red) for the question, ground-truth, and prediction.}
    \label{fig:llm_prompt}
\end{figure}
\subsubsection*{LLM-based Score}
Traditional metrics that rely on string-matching often fail to capture semantic correctness. They can penalize valid rephrasing and reward superficially similar but semantically incorrect predictions ~\cite{manas2024improving, bulian2022tomayto}. To overcome these limitations, we utilize an LLM-based Score, adopting the ``LLM-as-a-judge'' paradigm ~\cite{zheng2023judging, yu2023mm}, which demonstrates a higher correlation with human judgment ~\cite{gu2024survey, li2024generation}. Our method employs Qwen3-8B model~\cite{yang2025qwen3} to evaluate each prediction against its ground-truth answers. The evaluation is strictly guided by the instructions detailed in the prompt shown in Figure~\ref{fig:llm_prompt}. These rules specify how to handle various answer formats and enforce a structured output for robust parsing. This process yields the number of matched sub-answers ($m_i$) out of the total ($t_i$). The score for a single sample $i$ is the ratio of matched to total sub-answers:
$S_i = \frac{m_i}{t_i}$
For standard single-answer questions, $t_i$ is 1. The final LLM Score is the average across all $N$ samples in the dataset: $\text{LLM Score} = \frac{1}{N} \sum_{i=1}^{N} S_i$.

\subsection{Experimental Setups}
All experiments use the public DAM-3B checkpoint in half-precision on an NVIDIA RTX A5000. Text generation employs temperature \(1\times10^{-7}\), nucleus sampling \(p=0.5\), beam size 1, and dataset-specific token limits. Images are resized (longest side 1024 px) with a high-quality filter, and masks are single-channel grayscale. Other models follow their official implementation’s settings but share the DAM prompt and generation parameters.

\begin{table*}[t]
\centering
\scriptsize
\sisetup{
detect-weight,
mode=text,
table-format=2.2
}
\setlength{\tabcolsep}{2pt}
\resizebox{\textwidth}{!}{%
\begin{tabular}{
l
c
S[table-format=2.2] S[table-format=2.2]
S[table-format=2.2] S[table-format=2.2]
S[table-format=2.2] S[table-format=2.2]
S[table-format=2.2] S[table-format=2.2]
S[table-format=2.2] S[table-format=2.2]
S[table-format=2.2] S[table-format=2.2]
}
\toprule
\multirow{2}{*}{\textbf{Model}}
& \multirow{2}{*}{\shortstack{\textbf{Params} \\ \textbf{(in billions)}}}
& \multicolumn{2}{c}{\textbf{DocVQA}}
& \multicolumn{2}{c}{\textbf{InfographicVQA}}
& \multicolumn{2}{c}{\textbf{TextVQA}}
& \multicolumn{2}{c}{\textbf{ChartQA}}
& \multicolumn{2}{c}{\textbf{ChartQAPro}}
& \multicolumn{2}{c}{\textbf{VQAv2}} \\
\cmidrule(lr){3-4} \cmidrule(lr){5-6} \cmidrule(lr){7-8} \cmidrule(lr){9-10} \cmidrule(lr){11-12} \cmidrule(lr){13-14}
&
& {\bfseries ANLS} & {\bfseries LLM}
& {\bfseries ANLS} & {\bfseries LLM}
& {\bfseries VQAS} & {\bfseries LLM}
& {\bfseries RAcc} & {\bfseries LLM}
& {\bfseries RAcc} & {\bfseries LLM}
& {\bfseries VQAS} & {\bfseries LLM} \\
\midrule

\multicolumn{14}{l}{\textbf{Generalist VLMs}} \\
Phi-4-multimodal ~\cite{abouelenin2025phi} & 5B & 91.77 & 90.05 & 70.90 & 66.88 & 76.17 & 81.23 & \bfseries 84.44 & \bfseries 76.82 & 27.67 & 26.15 & 70.58 & 72.87 \\
LLaVA 1.6 Vicuna \cite{li2024llava}        & 7B & 58.84 & 51.39 & 20.37 & 18.92 & 56.53 & 61.87 & 52.68 & 42.40 & 16.21 & 20.99 & 70.59 & 72.35 \\
Molmo D ~\cite{deitke2024molmo}  & 7B & 82.78 & 81.35 & 61.45 & 58.03 & 74.77 & 81.39 & 57.40 & 65.06 & 28.51 & 28.71 & 65.56 & 72.81 \\
VideoLLaMA3 Image ~\cite{zhang2025videollama} & 7B & 91.73 & 89.97 & 69.47 & 65.43 & 62.06 & 67.18 & 67.40 & 63.46 & 28.62 & 31.00 & 57.85 & 60.22 \\
Qwen2.5-VL ~\cite{bai2025qwen2}  & 7B & \bfseries 94.04 & 92.84 & 77.93 & 73.88 & \bfseries 83.03 & \bfseries 87.41 & 80.04 & 75.90 & 34.72 & 32.40 & \bfseries 80.67 & \bfseries 82.63 \\
InternVL3 ~\cite{zhu2025internvl3}  & 8B & 92.05 & 90.09 & 71.80 & 66.55 & 78.63 & 82.55 & 77.44 & 71.32 & 29.24 & 30.87 & 74.90 & 76.98 \\
Ovis2 ~\cite{lu2024ovis}             & 8B & 93.77 & 91.61 & 78.96 & 74.89 & 82.13 & 86.83 & 82.88 & 76.14 & 32.65 & 36.47 & 76.85 & 79.17 \\
MiniCPM-o2.6 ~\cite{yao2024minicpm}  & 8B & 92.38 & 90.87 & 69.41 & 64.99 & 78.35 & 82.24 & 83.20 & 74.82 & 29.72 & 31.66 & 77.18 & 79.38 \\
Gemini-2.5-Flash~\cite{gemini25} &  -  & 93.20 & \bfseries 93.03 & \bfseries 85.32 & \bfseries 84.60 & 76.53 & 82.83 & 46.48 & 55.86 & \bfseries 49.99 & \bfseries 58.04 & 56.46 & 62.06 \\
\midrule
\multicolumn{14}{l}{\textbf{Region-aware VLMs}} \\
Shikra ~\cite{chen2023shikra}        & 7B &  5.56 &  6.26 &  8.07 & 8.55 & 26.13 & 31.19 &  8.20 &  6.20 &  3.54 & 10.26 & 51.37 & 62.68 \\
GPT4RoI ~\cite{zhang2025gpt4roi}     & 7B &  0.45 &  4.53 &  0.22 & 10.92 &  0.01 &  4.73 &  0.00 &  6.06 &  0.24 & 10.26 &  0.01 & 32.91 \\
Ferret ~\cite{you2023ferret}         & 7B &  0.66 &  4.13 &  0.19 & 5.71 &  0.43 &  5.21 &  0.04 &  3.25 &  0.45 &  8.44 &  0.02 &  5.48 \\
Osprey ~\cite{yuan2024osprey}        & 7B &  0.97 &  5.05 &  2.36 &  8.90 &  0.34 & 14.13 &  1.56 &  5.38 &  0.22 & 11.88 &  6.72 & 43.37 \\
ViP-LLaVA ~\cite{cai2024vip}         & 7B & 22.35 & 15.44 & \bfseries 22.31 & \bfseries 17.28 & 47.35 & 52.59 & 16.72 & 11.09 & 11.88 & 15.43 & 76.55 & 79.16 \\
PAM ~\cite{lin2025perceive}          & 3B &  0.26 &  3.53 &  0.89 & 4.37 &  0.42 &  5.57 &  0.00 &  1.56 &  0.00 &  4.89 &  0.65 & 16.05 \\
DAM ~\cite{lian2025describe}         & 3B & 35.22 & 28.48 & 19.27 & 14.91 & 57.86 & 65.22 & 46.52 & 35.62 & \bfseries 18.90 & \bfseries 21.81 & \bfseries 79.25 & \bfseries 82.57 \\
DAM-QA (Ours)        & 3B & \bfseries 42.34 & \bfseries 35.60 & 20.25 & 16.12 & \bfseries 59.67 & \bfseries 67.29 & \bfseries 47.72 & \bfseries 40.06 & 14.88 & 18.98 & 79.20 & 82.51 \\
\bottomrule
\end{tabular}%
}
\caption{Performance comparison on six VQA benchmarks across generalist and region-aware Vision-Language Models (VLMs). Region-aware models tend to be less competitive than generalist VLMs in overall performance. However, the proposed DAM-QA, which incorporates a simple sliding-window strategy into the DAM input pipeline, consistently improves upon the baseline DAM and outperforms all existing region-aware models across nearly all benchmarks.}
\label{tab:full_vqa_results}
\end{table*}

\subsection{Result}
\label{sec:main_exp}
In this section, we conduct extensive experiments on our proposed DAM-QA alongside VLMs belonging to two distinct categories, aiming to establish a comprehensive performance benchmark for the task of text-rich image VQA. These categories correspond to generalist and region-aware models, differentiated by their design objectives and approach to processing visual inputs.

\textbf{Generalist VLMs}: This group includes current state-of-the-art models: Phi-4 Multimodal, LLaVA 1.6 Vicuna, Molmo, VideoLLaMA3 Image, Qwen2.5-VL, InternVL3, Ovis2, MiniCPM-o2.6, and Gemini-2.5-Flash.

\textbf{Region-aware VLMs}: This category comprises models specifically designed to process designated image regions such as Shikra, GPT4RoI, Ferret, Osprey, and ViP-LLaVA.

To guarantee a fair and controlled comparison, both generalist and region-aware models are evaluated using a consistent prompt template detailed in Section~\ref{sec:prompt} across all datasets, regardless of model type. The results in Table~\ref{tab:full_vqa_results} reveal a significant performance disparity between generalist and region-aware VLMs. Top-tier generalist models, particularly Qwen2.5-VL and Gemini-2.5-Flash, set a high performance bar on most benchmarks, demonstrating robust capabilities in holistic understanding. In contrast, Shikra, GPT4RoI, and Ferret struggle significantly on text-intensive datasets. This suggests that existing region-aware specialization does not inherently guarantee strong performance on these complex VQA tasks. An exception is ViP-LLaVA, which demonstrates competitive performance on specific benchmarks like InfographicVQA and VQAv2. Against this backdrop, DAM-QA stands out among region-aware models, surpassing all evaluated peers on five of six benchmarks. While not universal SOTA, DAM-QA demonstrates consistently strong performance across diverse VQA tasks.



In summary, DAM-QA achieves highly competitive performance among region-aware methods, establishing a stronger and more reliable baseline for future research in region-focused VQA.

\subsection{Ablation Study}
We conduct a series of targeted ablation studies to evaluate the impact of key design choices in DAM-QA, thus providing empirical justification for our proposed methodology. Specifically, we analyze our prompt rules for visual grounding and answer abstention (Section~\ref{sec:ab_study_prompt_design}), determine the optimal sliding-window parameters (Section~\ref{sec:ab_study_patch_generation}), and validate the voting strategy for handling ``unanswerable'' patch-level predictions (Section~\ref{sec:ab_study_voting}).

\subsubsection{Effect of VQA Prompt Design on DAM-QA}
\label{sec:ab_study_prompt_design}

To understand how each component of our VQA prompt contributes to performance, we isolate the two core rules (faithful grounding and abstention) and measure their effects both individually and in combination. Three benchmarks were selected to stress different aspects of text-centric VQA: DocVQA for document layout understanding, TextVQA for reading text in natural scenes, and ChartQAPro for handling multi-chart reasoning and unanswerable cases. The unified prompt (Section~\ref{sec:prompt}) comprises two core rules:

\begin{enumerate}
    \item VQA rule 1 (Image‐Only Justification) forces DAM to ground every answer solely in the pixels of the input image, thereby reducing hallucinations.
    \item VQA rule 2 (Abstention) instructs DAM to respond with ``unanswerable'' when the image alone lacks sufficient information, preventing unsupported guesses.
\end{enumerate}

\begin{table}[!ht]
\centering
\scriptsize
\sisetup{
  detect-weight,
  mode=text,
  table-format=2.2
}
\setlength{\tabcolsep}{2pt}
\resizebox{\columnwidth}{!}{%
\begin{tabular}{
  l 
  cc 
  S S 
  S S 
  S S
}
\toprule
\multirow{2}{*}{\textbf{Model}}
  & \multicolumn{2}{c}{\textbf{Prompt Variant}}
  & \multicolumn{2}{c}{\textbf{DocVQA}}
  & \multicolumn{2}{c}{\textbf{TextVQA}}
  & \multicolumn{2}{c}{\textbf{ChartQAPro}} \\ 
\cmidrule(lr){2-3} \cmidrule(lr){4-5} \cmidrule(lr){6-7} \cmidrule(lr){8-9}
& \textbf{Rule 1} & \textbf{Rule 2} 
& {\bfseries ANLS} & {\bfseries LLM} 
& {\bfseries VQAS} & {\bfseries LLM} 
& {\bfseries RAcc}   & {\bfseries LLM} \\
\midrule
DAM~\cite{lian2025describe} 
  & \checkmark & \checkmark 
  & 34.84 & 25.44 
  & 57.86 & 65.22 
  & \textbf{18.90} & \textbf{21.81} \\
\midrule
\multirow{4}{*}{DAM-QA}
  & – & – 
  & 39.83 & 29.99 
  & 58.17 & 63.65 
  & 9.28 & 12.56 \\

  & \checkmark & – 
  & 39.49 & 29.76 
  & 58.17 & 63.41 
  & 9.77 & 13.49 \\

  & – & \checkmark 
  & 41.02 & 31.02 
  & 59.27 & 64.56 
  & 16.75 & 17.13 \\

  & \checkmark & \checkmark 
  & \textbf{42.34} & \textbf{32.18} 
  & \textbf{59.67} & \textbf{67.29} 
  & 14.88 & 15.65 \\
\bottomrule
\end{tabular}%
}
\caption{Ablation of prompt variants for VQA. Rule 1 enforces image-only justification; Rule 2 enables abstention when there is insufficient evidence.}
\label{tab:ablation_prompt}
\end{table}

Table~\ref{tab:ablation_prompt} shows that combining both prompt rules yields the best overall performance. Using only the image-justification rule reduces hallucinations but leads to unsupported answers, hurting scores, especially on ChartQAPro. Conversely, the abstention rule prevents unsupported guesses, improving results on DocVQA, TextVQA, and partially recovering performance on ChartQAPro. However, each rule alone underperforms the combined setup, which best balances grounding and abstention, addressing both hallucination and overcommitment across diverse VQA tasks.

\subsubsection{Patch Generation Parameters}
\label{sec:ab_study_patch_generation}

\begin{table}[!ht]
\centering
\scriptsize
\sisetup{detect-weight, mode=text, table-format=2.2}
\setlength{\tabcolsep}{2pt}
\renewcommand{\arraystretch}{1.15}
\resizebox{\columnwidth}{!}{%
\begin{tabular}{
  l    
  c    
  c    
  S S  
  S S  
  S S  
}
\toprule
\textbf{Model} & \textbf{Win} & \textbf{Stride}
  & \multicolumn{2}{c}{\textbf{DocVQA}} 
  & \multicolumn{2}{c}{\textbf{TextVQA}} 
  & \multicolumn{2}{c}{\textbf{ChartQA}} \\
\cmidrule(lr){4-5} \cmidrule(lr){6-7} \cmidrule(lr){8-9}
& & & \textbf{ANLS} & \textbf{LLM} & \textbf{VQAS} & \textbf{LLM} & \textbf{RAcc} & \textbf{LLM} \\
\midrule
DAM~\cite{lian2025describe} & – & – 
  & 35.22 & 28.48 & 57.86 & 65.22 & 46.52 & 35.62 \\
\midrule
\multirow{3}{*}{\makecell[l]{DAM-QA\\(Granularity)}} 
  & 256 & 128 
  & 36.02 & 27.17 & 45.31 & 49.82 & 41.32 & 31.00 \\
& 512 & 256 
  & 42.34 & 32.18 & \textbf{59.67} & \textbf{64.93} & \textbf{47.72} & \textbf{39.16} \\
& 768 & 384 
  & 39.35 & 29.90 & 59.45 & 64.43 & 46.64 & 34.82 \\
\midrule
\multirow{2}{*}{\makecell[l]{DAM-QA\\(Sweep Window)}} 
  & 256 & 256 
  & 39.78 & 29.87 & 58.34 & 63.50 & 46.00 & 34.28 \\
& 768 & 256 
  & 39.89 & 30.80 & 59.51 & 64.43 & 46.64 & 34.76 \\
\midrule
\multirow{2}{*}{\makecell[l]{DAM-QA\\(Sweep Stride)}} 
  & 512 & 128 
  & \textbf{43.22} & \textbf{35.79} & 59.31 & 64.27 & 46.56 & 38.26 \\
& 512 & 384 
  & 41.62 & 30.89 & 59.13 & 64.27 & 45.36 & 35.58 \\
\bottomrule
\end{tabular}%
}
\caption{Ablation on patch generation parameters. Each configuration belongs to a category: Granularity, Sweep Window, or Sweep Stride, as shown in parentheses. All runs on DAM-QA use the sliding-window extension and full prompt (both VQA rules).}
\label{tab:ablation_patch}
\end{table}

Section~\ref{sec:sliding} introduces our sliding‐window protocol, and Table~\ref{tab:ablation_patch} evaluates its key parameters on DocVQA, TextVQA and ChartQA. We vary the patch size and stride to analyze granularity effects. Compared to the $512\times256$ default, $256$‐pixel/128‐pixel patches cause official metrics (ANLS, VQA Score, RAcc) to drop by 15\%, 24\% and 13\%, and LLM scores to drop by 16\%, 23\% and 21\%, while $768$‐pixel/384‐pixel patches cause smaller drops of 7\%, 0.4\% and 2\% in official metrics and of 7\%, 1\% and 11\% in LLM scores. Fixing stride at $256$ pixels gives drops of about 6\%, 2\% and 4\% in official metrics and of 7\%, 2\% and 12\% in LLM, while changing stride with a $512$‐pixel window produces only minor shifts in both official and semantic scores. These results show that $512\times256$ strikes the optimal balance between fine detail and global context.

\begin{figure}[t]
    \centering
    \includegraphics[width=0.95\linewidth]{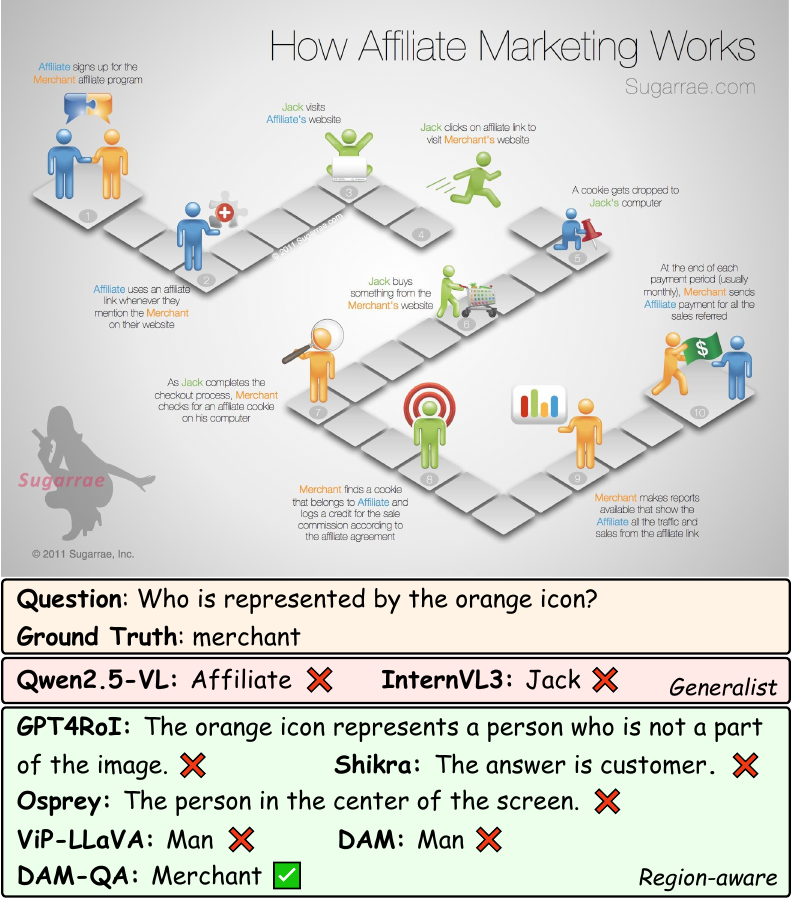} 
    \caption{Qualitative results on an InfographicVQA~\cite{mathew2022infographicvqa} sample requiring precise visual grounding. Thanks to the proposed mechanism, DAM-QA can successfully answer the question that technically needs to read the word ``Merchant'' inside the image.}
    \label{fig:quantitative}
\end{figure}

\subsubsection{Vote Weight}
\label{sec:ab_study_voting}

\begin{table}[!ht]
\centering
\scriptsize
\sisetup{
  detect-weight,
  mode=text,
  table-format=2.2
}
\setlength{\tabcolsep}{2pt}
\resizebox{\columnwidth}{!}{%
\begin{tabular}{
  l    
  c    
  S[table-format=2.2] S[table-format=2.2]  
  S[table-format=2.2] S[table-format=2.2]  
  S[table-format=2.2] S[table-format=2.2]  
}
\toprule
\multirow{2}{*}{\textbf{Model}}
  & \multirow{2}{*}{\textbf{Weight}}
  & \multicolumn{2}{c}{\textbf{DocVQA}}
  & \multicolumn{2}{c}{\textbf{TextVQA}}
  & \multicolumn{2}{c}{\textbf{VQAv2}} \\
\cmidrule(lr){3-4} \cmidrule(lr){5-6} \cmidrule(lr){7-8}
& 
& {\bfseries ANLS} & {\bfseries LLM}
& {\bfseries VQAS} & {\bfseries LLM}
& {\bfseries VQAS} & {\bfseries LLM} \\
\midrule
DAM~\cite{lian2025describe} & –  
  & 34.84 & 25.44 
  & 57.84 & 62.91 
  & \textbf{79.25} & \textbf{81.46} \\
\midrule
\multirow{4}{*}{DAM-QA}
  & 0.0 
  & \textbf{42.34} & \textbf{32.18} 
  & \textbf{59.67} & \textbf{64.93} 
  & 79.20 & 81.40 \\

  & 0.5 
  & 33.76 & 26.24 
  & 58.96 & 64.03 
  & 79.24 & \textbf{81.46} \\

  & 1.0 
  & 23.63 & 19.41 
  & 53.91 & 58.01 
  & 79.20 & 81.40 \\

  & 1.5 
  & 17.82 & 14.86 
  & 47.68 & 51.03 
  & 79.16 & 81.36 \\
\bottomrule
\end{tabular}%
}
\caption{Ablation of the ``unanswerable'' vote weight. The Weight column denotes the vote assigned to patch predictions labeled ``unanswerable''. All other settings follow the default sliding-window extension (~\ref{sec:sliding}) and prompt (\ref{sec:prompt}).}
\label{tab:ablation_weight}
\end{table}

\noindent Despite Section~\ref{sec:ab_study_prompt_design}, which showed that abstention reduces unsupported guesses, we hypothesize that patches labeled “unanswerable” add noise when treated like regular answers. Therefore, in Section~\ref{sec:weighted_voting_scheme}, we set the vote weight for “unanswerable” predictions to $0.0$. To validate this choice, we vary the vote weight (from $0.0$ to $1.5$) assigned to “unanswerable” patch predictions in our ablation study on three benchmarks: DocVQA, TextVQA, and VQAv2. Table~\ref{tab:ablation_weight} reports both official metrics (ANLS or VQA Score) and LLM semantic scores. Performance consistently peaks at weight $0.0$. Assigning a small nonzero weight (0.5) causes DocVQA ANLS to fall by $20\%$ and its LLM score by $18\%$, while TextVQA VQA Score and LLM score each drop by $1\%$. VQAv2 changes by less than $0.1\%$. These results support zero‐weighting abstentions as the best overall choice.

\section{Conclusion}
\label{sec:conclusion}
In this work, we addressed Visual Question Answering (VQA) on text-rich images using region-aware Vision-Language Models (VLMs), proposing DAM-QA, an effective adaptation of the Describe Anything Model (DAM). Our framework integrates a sliding-window protocol and multi-scale aggregation to enhance DAM’s ability to reason over localized visual content. Extensive evaluations across multiple benchmarks show that DAM-QA consistently outperforms the baseline DAM and all existing region-aware models, while remaining competitive with larger generalist models, despite DAM’s smaller size (3B vs 7B parameters). These results show that descriptive models, such as DAM, can excel in analytical, text-centric visual tasks. Future work may explore fine-tuning strategies for region-aware VQA.
\section{Acknowledgement}
This work was supported in part by U.S. NSF grant \\MCB-2205148.
{
    \small
    \bibliographystyle{ieeenat_fullname}
    \bibliography{main}
}

\end{document}